%% file: ms.tex
\documentclass[10pt,twocolumn,letterpaper]{article}

\usepackage{iccv}
\usepackage{times}
\usepackage{epsfig}
\usepackage{xcolor}
\usepackage{graphicx}
\usepackage{amsmath}
\usepackage{amssymb}
\usepackage{bbm}
\usepackage{textcomp}
\usepackage{booktabs}
\usepackage{microtype}

\usepackage[export]{adjustbox}
\usepackage{mleftright}
\mleftright

\usepackage[pagebackref=true,breaklinks=true,letterpaper=true,colorlinks=false,bookmarks=false]{hyperref}

\usepackage[linesnumbered,ruled]{algorithm2e}

\iccvfinalcopy %

\def\iccvPaperID{12} %

\input{defs}

\ificcvfinal\fi
\begin{document}

\title{Leveraging Vision Reconstruction Pipelines for Satellite Imagery}

\author{Kai Zhang, Jin Sun, Noah Snavely\\
{Cornell Tech, Cornell University}\\
}

\maketitle
\thispagestyle{empty}

\begin{abstract}
\input{00-abstract}
\end{abstract}

\input{01-intro.tex}
\begin{figure*}[t]
\centering
\includegraphics[width=0.78\textwidth]{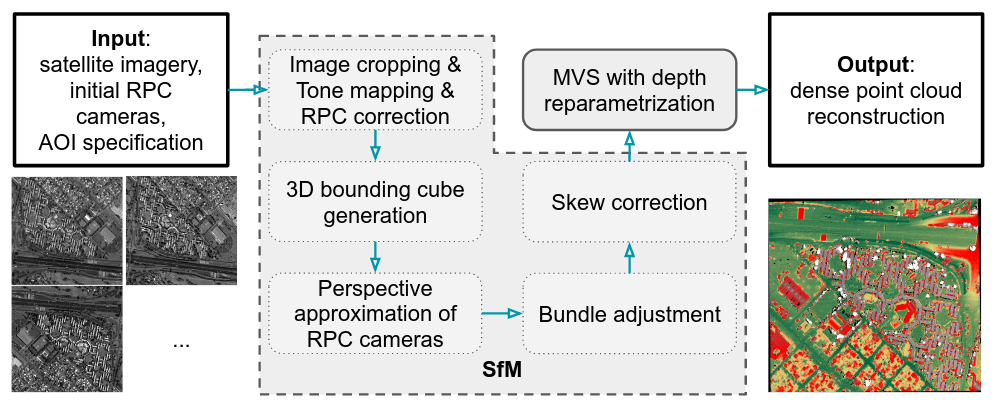}
\caption{Our proposed satellite-adapted 3D reconstruction pipeline. Dashed borders indicate SfM modules and solid borders indicate MVS. Point cloud is displayed as height map on a geographic grid, with color scale in Fig.~\ref{fig:main_qualitative}.}
\label{fig:full_pipeline}
\end{figure*}

\input{02-related.tex}
\input{03-method_sfm.tex}

\input{04-method_mvs.tex}

\input{05-experiments.tex}

\input{06-conclusion.tex}

\medskip
{\small 
\noindent \textbf{Acknowledgements.} The research is based upon work supported by the Office of the Director of National Intelligence (ODNI), Intelligence Advanced Research Projects Activity (IARPA), via DOI/IBC Contract Number D17PC00287. The views and conclusions contained herein are those of the authors and should not be interpreted as necessarily representing the official policies or endorsements, either expressed or implied, of the ODNI, IARPA, or the U.S.\ Government. The U.S.\ Government is authorized to reproduce and distribute reprints for Governmental purposes notwithstanding any copyright annotation thereon. We thank Zhengqi Li and Charles Herrmann for helpful discussions.}

{\small
\bibliographystyle{ieee}
\bibliography{egbib}
}

\end{document}

%% file: defs.tex
\newenvironment{packed_item}{
\begin{itemize}
\vspace{0pt}
  \setlength{\itemsep}{5pt}
  \setlength{\parskip}{0pt}
  \setlength{\parsep}{0pt}
  \setlength{\topsep}{-10pt}
  \setlength{\partopsep}{0pt}
}{\end{itemize}}

%% file: 00-abstract.tex
Reconstructing 3D geometry from satellite imagery is an important topic of research. However, disparities exist between how this 3D reconstruction problem is handled in the remote sensing context and how multi-view reconstruction pipelines have been developed in the computer vision community. In this paper, we explore whether state-of-the-art reconstruction pipelines from the vision community can be applied to the satellite imagery. Along the way, we address several challenges adapting vision-based structure from motion and multi-view stereo methods.
We show that vision pipelines can offer competitive speed and accuracy in the satellite context.

%% file: 01-intro.tex
\section{Introduction}
Satellite imagery finds applications in a variety of domains, including ecological monitoring, 3D urban modelling, and navigation.
The growing number of academic and commercial satellites, and increasing accessibility of satellite imagery, has sparked greater interest in 
use of satellite imagery for large-scale 3D reconstruction of the Earth's surface~\cite{kuschk2013large,facciolo2017automatic,bosch2016multiple,bosch2017metric}.
Satellites can capture the same geographic area over the course of seconds, days, months, and years, yielding a wealth of %
imagery available for 3D reconstruction.

Much of the research on 3D reconstruction from satellite imagery has progressed in the remote sensing community. At the same time, the computer vision community has also seen great advances in 3D reconstruction, primarily using ground-level views of objects, buildings, and interiors~\cite{snavely2006photo,snavely2008modeling,agarwal2009building,schonberger2016structure}. As a result, several software packages for high-quality reconstruction are available~\cite{jancosek2011multi,fuhrmann2014mve,schonberger2016structure,openMVG}. While developments in the remote sensing and computer vision communities have proceeded largely independently over the past few decades, it seems natural that advances in one area should benefit the other, since the fundamental challenges involved are similar. However, the solutions produced in each community have diverged in several ways:
\begin{packed_item}
  \item In remote sensing, recent satellite-based reconstruction methods often focus on the minimal \emph{two-view}  scenario~\cite{de2014automatic2,beyer2018ames,rupnik2017micmac} within a paradigm of fusing multiple two-view reconstructions. This strategy works well empirically~\cite{toutin2004geometric}, e.g., in the state-of-the-art S2P satellite reconstruction pipeline~\cite{facciolo2017automatic}. In contrast, full multi-view stereo methods have long been explored in computer vision due to their advantages in areas of occlusion, repeated patterns, low image signal, etc.
  \item While the full set of satellite images available for a certain location might span years, remote sensing methods tend to conservatively select pairs of images within much shorter time span for reconstruction due to the associated diversity in illumination, season, weather, etc. In contrast, recent vision pipelines such as COLMAP~\cite{schonberger2016structure, schonberger2016pixelwise}, are designed for Internet photos captured in the wild under highly diverse conditions.
  \item Accordingly, methods like COLMAP focus on scalability to thousands or millions of views, and hence strive for efficient algorithms suitable for GPU computing.
  \item Finally, there are significant differences in terminology and assumptions between communities. For example, satellite cameras are typically described by complex black-box models, while computer vision pipelines assume simpler pinhole cameras. 
\end{packed_item}

In this paper, we explore the question of whether these gaps can be bridged, and in particular, whether state-of-the-art reconstruction pipelines developed in the vision community can be leveraged for satellite reconstruction problems. If yes, then the satellite stereo problem can benefit from progress 
made in the computer vision community.

Specifically, we take COLMAP, a computer vision pipeline that performs both structure from motion (SfM) and multi-view stereo (MVS), and a top performer in recent ground-level reconstruction benchmarks~\cite{knapitsch2017tanks}, and adapt it to the satellite image scenario. Along the way, we address a number of key technical challenges, described in this paper. Our adapted stereo pipeline is much more efficient compared to S2P, and can indeed operate on large numbers of diverse images. This method achieves sub-meter accuracy, although its accuracy is lower compared to S2P, in part due to its prioritization of efficiency during the MVS stage. Hence, we also develop a simple plane-sweep stereo alternative that performs comparably to S2P in terms of accuracy.

Our adapted reconstruction pipeline is shown in Fig.~\ref{fig:full_pipeline}, which highlights some key enhancements to COLMAP. The same adaptation techniques can also be applied to other MVS pipelines. In particular, the challenges we faced include:

\smallskip
\noindent \textbf{For SfM:} a main difficulty arises from the complex Rational Polynomial Camera (RPC) model~\cite{hartley1997cubic,tao2001comprehensive,hu2004understanding} that is widely used for satellite images. 
We propose a method to reduce the RPC model locally to a simpler pinhole camera model, which in turn allows us to utilize state-of-the-art computer vision SfM pipelines~\cite{schonberger2016structure} to bundle adjust the camera parameters.

\smallskip
\noindent \textbf{For MVS:} a major issue involves the lack of numerical precision due to depths that concentrate in an interval that is far from the sensor plane, as a result of the large distance between satellites and the Earth. Numerical precision issues are particular severe on consumer GPUs, which are limited to single-precision floating point computation. Our solution is to re-parameterize depth using plane-plus-parallax~\cite{irani1998from}.

In addition to these contributions, we also have released our code 
implementing these changes.
We hope that this software can help make it easier for 3D computer vision researchers to explore the growing domain of satellite imagery.

%% file: 02-related.tex
\section{Background and related work}

\noindent \textbf{Satellite images.}
Research dating back to the early 2000s \cite{tao2001comprehensive,hu2004understanding} has promoted the RPC camera model as a generic replacement of a physical sensor model for 
geo-mapping.
The RPC model commonly used includes 78 coefficients and 10 normalization constants:
\begin{align}
    u&=\mu_u+\sigma_u\cdot g\left(\frac{x-\mu_x}{\sigma_x},\frac{y-\mu_y}{\sigma_y},\frac{z-\mu_z}{\sigma_z}\right) \\
    v&=\mu_v+\sigma_v\cdot h\left(\frac{x-\mu_x}{\sigma_x},\frac{y-\mu_y}{\sigma_y},\frac{z-\mu_z}{\sigma_z}\right)
,\end{align}
where $x, y, z$ denote latitude, longitude, and altitude,
$g, h$ are ratios of two cubic polynomials parameterized by 39 coefficients, and $\mu_{\{x,y,z,u,v\}}$ and $\sigma_{\{x,y,z,u,v\}}$
are normalization constants. While lacking physical interpretability, the RPC model has been 
successful in downstream photogrammetry tasks such as ortho-photo generation and DEM extraction~\cite{hu2004understanding}. It has consequently become a standard practice over the past decade for satellite image vendors to deliver an RPC model to accompany each satellite image.

Based on the RPC model, prior work has explored parameter estimation, bundle adjustment, and various minimal solvers. Parameter estimation~\cite{hartley1997cubic} requires knowledge of the physical sensor, which is often inaccessible to end users, and so such estimation is generally performed by the satellite image vendors. However, efforts have sought to understand the residual error in vendor-provided RPCs, leading to the notion of the bias-corrected RPC model~\cite{fraser2005bias}. This work observes that the RPC model can be subject to a bias equivalent to a 2D image-space translation, and this bias can be corrected with the aid of human surveyed ground control points. Grodecki and Dial~\cite{grodecki2003block} propose an adjustable image-space translation term to facilitate bundle adjustment of multiple RPCs. Later, de Franchis et al.~\cite{de2014automatic} propose a similar method to fix relative pointing error in a stereo pair. Regarding minimal solvers, inverse projection and triangulation are critical for geometry-related tasks. For RPC models, they involve solving non-trivial third-order polynomial systems~\cite{zheng2015minimal}.

In the remote sensing community, multiple satellite stereo pipelines 
are based on the RPC model~\cite{de2014automatic2,beyer2018ames,rupnik2017micmac,shean2016automated}. These often focus on a minimal two-view stereo setting rather than utilizing multiple views at once. The two-view stereo components rely on  disparity~\cite{de2014automatic2,rupnik2017micmac,beyer2018ames} or optical flow estimation~\cite{bosch2016multiple} to find dense correspondence, followed by RPC-based triangulation. A representative recent work, S2P~\cite{facciolo2017automatic}, heuristically ranks all pairs of images in an input set, then aligns and merges independent reconstructions produced via two-view stereo on the top 50 image pairs~\cite{de2014automatic2}. We compare our approach primarily to S2P, as it was the winner of the IARPA Multi-View Stereo 3D Mapping Challenge~\cite{bosch2016multiple}.

\medskip
\noindent \textbf{Ground-level images.}
Pipelines developed in the computer vision community adopt the easy-to-interpret pinhole camera model, in contrast to the RPC model. Using homogeneous coordinates, the pinhole model is written as\footnotemark
\begin{equation}
[u; v; 1]=\mathbf{K}[\mathbf{R}\  \mathbf{t}][x;y;z;1]
\end{equation}\footnotetext{In this paper, we follow MATLAB notation for denoting column vectors, row vectors, and matrices. Vectors and matrices are in boldface.}
The intrinsics matrix $\mathbf{K}\in \mathcal{R}^{3\times 3}$ takes the form $[f_x, s, c_x; 0, f_y, c_y; 0, 0, 1]$, while the extrinsics consist of a camera rotation $\mathbf{R} \in \mathrm{SO}(3)$ and translation $\mathbf{t}\in \mathcal{R}^{3}$. In many scenarios, one can assume zero skew ($s=0$), unit aspect ratio ($f_x=f_y$), and centered principal point ($c_x,c_y=0.5$ in normalized image coordinates).

Most modern computer vision reconstruction pipelines consist of SfM and MVS stages. SfM aims to obtain accurate camera parameters by iteratively optimizing these parameters along with a triangulated sparse point cloud. We refer the readers to~\cite{schonberger2016structure} for recent progress in SfM. MVS reconstructs a dense point cloud or mesh given a set of images and their corresponding camera parameters. 
Key issues in MVS include view selection based on camera baseline, overlap, etc, and robust recovery of dense depth maps~\cite{schonberger2016pixelwise}. From a computational perspective, many MVS pipelines can scale to thousands of images thanks to the use of GPU hardware.

%% file: 03-method_sfm.tex
\section{Adapting SfM to satellite imagery}
The goal of SfM is to recover accurate camera parameters for use in subsequent steps such as dense reconstruction. A major factor distinguishing satellite stereo pipelines from typical vision pipelines is the camera model (RPC vs.\ pinhole). In this section, we bridge this gap by locally approximating the RPC model with a pinhole model. Following this approximation, modern SfM pipelines, e.g., VisualSFM~\cite{wu2013towards} and COLMAP SfM~\cite{schonberger2016structure}
can be utilized to bundle adjust the camera parameters.

\subsection{Coordinate system} \label{sec:coordinate system}
Our Earth is approximately ellipsoidal. To locate a point on the Earth's surface, the RPC model uses a global coordinate frame consisting of (latitude, longitude, altitude). This coordinate frame is defined with respect to a nominal reference ellipsoid, e.g., World Geodetic System 1984 (WGS84).

The (latitude, longitude, altitude) coordinate system is not a Cartesian frame as assumed by most vision pipelines. In real-world applications, however, we almost always focus on a much smaller geometric area, e.g.,  $1\ \text{km}^2$, either because our reconstruction task is regional or a large area is divided into smaller patches for parallel processing. Hence, we adopt a simpler local Cartesian coordinate system for a specific reconstruction problem. 
In particular, we use the East-North-Up (ENU) coordinate system, defined by first choosing an observer point $(\mathit{latitude}_0, \mathit{longitude}_0, \mathit{altitude}_0)$, and then the ``east'', ``north'', and ``up'' directions at this point form the three axes. The ``east-north'' plane is parallel to the tangent plane  of the reference ellipsoid at the point $(\mathit{latitude}_0, \mathit{longitude}_0, 0)$. With a careful choice of the observer point, 
such a local Cartesian frame not only opens the door to a pinhole model approximation, but also improves numerical stability by bringing the scene close to the origin. 

\subsection{3D bounding cube generation}
In the (latitude, longitude, altitude) coordinate system, an area of interest (AOI) is specified 
by a 2D bounding box $[x_{\textit{min}}, x_{\textit{max}}]\times[y_{\textit{min}}, y_{\textit{max}}]$. To bound the AOI along the $z$-axis, we use public SRTM data describing the global terrain altitude up to an accuracy of 30m~\cite{jarvis2008hole}, together with the assumption that most vegetation and buildings are less than a few hundred meters in height.
We assume that the input satellite images are cropped to this AOI.

\subsection{Tonemapping}
Satellite images typically have high dynamic range (HDR). To accelerate computation, the input HDR satellite images are tonemapped to LDR. We observe that the original pixel intensities exhibit a long-tailed distribution, hence direct scaling results in dark, low-contrast LDR images.
Hence, we instead use a standard tonemapping operation consisting of a gamma correction $I_\mathrm{out} = I_\mathrm{in} ^ {1/2.2}$ (where $I$ denotes pixel intensity) followed by a scale factor.

\subsection{Perspective camera approximation}
We now mathematically justify the practice of locally approximating an RPC camera with a perspective camera,
then provide a numerical method to estimate camera parameters.

\medskip
\noindent \textbf{Mathematical justification.}
The physical satellite imaging process can be modelled by an orbiting linear pushbroom camera, such that each row of a satellite image
is captured at a slightly different time instant. For satellite images, the depth values $Z$ of scene points lie in the interval $ [\bar{Z}-\hat{Z}/{2},\bar{Z}-\hat{Z}/{2}]$, with $\bar{Z} \gg \hat{Z}$. Under this condition, we show that both perspective cameras and linear pushbroom cameras can be accurately approximated by the so-called weak perspective camera, and consequently we can approximately transform a linear pushbroom camera into a perspective camera.

First, as a result of small scene depth variation and very large average scene depth, a perspective camera reduces to a weak perspective camera that takes the form
$u\approx\frac{f_x}{\bar{Z}}X+\frac{s}{\bar{Z}}Y+c_x,
 v\approx\frac{f_y}{\bar{Z}}Y+c_y$, 
where $(X,Y,Z)$ is a point in the camera coordinate frame~\cite{forsyth2003modern}.\footnote{In this paper $(x, y, z)$ refers to a 3D point in the scene coordinate frame, while $(X,Y,Z)$ is a 3D point in a given camera coordinate frame.}

Next, we show that a linear pushbroom camera can also be reduced to a weak perspective camera under the same condition. Projection under a linear pushbroom camera can be modeled as: $
    u=a_1x+b_1y+c_1z + d_1,
    v=\frac{a_2x+b_2y+c_2z+d_2}{a_3x+b_3y+c_3z+d_3}$, where $(x,y,z)$ is a point in the scene coordinate frame.
In other words, satellite images exhibit linearity along the row axis $u$, while having a (weak) perspective effect along the column axis $v$. Let $f, \tilde{c}_x, \tilde{c}_y, \mathbf{R}, \mathbf{t}$ be the intrinsics and extrinsics (assuming zero skew and unit aspect ratio) of the perspective camera along the column axis. We then use $\mathbf{R},\mathbf{t}$ to transform scene coordinates $(x,y,z)$ to camera coordinates $(X,Y,Z)$, and rewrite the linear push-broom camera model. With some approximation, the result is again a weak perspective camera:
\begin{align}
    u&=\hat{a}_1X+\hat{b}_1Y+\hat{c}_1Z + \hat{d}_1 \nonumber\\
    &=\hat{a}_1X+\hat{b}_1Y+\hat{c}_1\bar{Z}\cdot \left(1+\frac{Z-\bar{Z}}{\bar{Z}}\right) + \hat{d}_1 \nonumber\\
    &\approx \hat{a}_1X+\hat{b}_1Y+\hat{c}_1\bar{Z}+ \hat{d}_1 \label{eq:linear_pushroom_u} \\
    v&=f\frac{Y}{Z}+\tilde{c}_y \label{eq:linear_pushroom_v}
\end{align}
Comparing the two weak perspective cameras, we have $
    \hat{a}_1=\frac{f_x}{\bar{Z}},
    \hat{b}_1=\frac{s}{\bar{Z}},
    \hat{c}_1\bar{Z}+\hat{d}_1=c_x,
    f=f_y,
    \tilde{c}_y=c_y$. Solving for $f_x, f_y,s, c_x, c_y$, together with $\mathbf{R},\mathbf{t}$, gives us the perspective camera approximation. Note that Eq.~\ref{eq:linear_pushroom_u}, \ref{eq:linear_pushroom_v} assumes that the satellite moves at a constant orientation and velocity, which is approximately true because of the extremely short time period needed to capture a small local area.

\medskip
\noindent \textbf{Numerical solution.}
The derivation above only establishes the existence of perspective approximations to satellite cameras in a local area. %
To actually solve for an approximated perspective camera model, we perform three steps: sampling the RPC model, fitting a projection matrix, and factorizing into a standard form.

We first sample the RPC model by generating a set of 3D-2D correspondences. Specifically, we uniformly discretize the ENU 3D bounding cube into a finite grid, with each axis evenly divided into $M$ sample locations. The resulting $M^3$ grid samples $(x,y,z)$ are then converted from ENU coordinates to (latitude, longitude, altitude), and projected via the RPC model into pixel coordinates $(u,v)$; this gives us a total of $M^3$ 3D-2D correspondences $(x_i,y_i,z_i,u_i,v_i),i=1,\dots,M^3$. 
We filter out correspondences whose pixel coordinates lie outside the image boundary.

Next, we solve for a $3\times 4$ projection matrix $\mathbf{P}=[\mathbf{p}_1^T; \mathbf{p}_2^T; \mathbf{p}_3^T]$. For simplicity, we denote $\mathbf{x}=[x;y;z;1]$, and then the projection equation of our perspective camera becomes $u=\mathbf{p}_1^T\mathbf{x}/\mathbf{p}_3^T\mathbf{x}, v=\mathbf{p}_2^T\mathbf{x}/\mathbf{p}_3^T\mathbf{x}$. Given a total of $L$ 3D-2D correspondences, $(\mathbf{x}_i,u_i,v_i),i=1,\dots,L$, we use the standard direct linear transformation method to solve for $\mathbf{P}$~\cite{Hartley2004}.
Finally, we factor $\mathbf{P}$ into the standard form $\mathbf{K}[\mathbf{R}\ \mathbf{t}]$~\cite{Hartley2004}. Note that this numerical solution outputs a full intrinsic matrix with a non-zero skew that is unusual for ground-level images, but can be explained in the satellite case by the equation $s=b_1\bar{Z}$ in our derivation above. The physical intuition is that satellite images are produced by stitching image rows captured at slightly different instants; hence rows may not be perfectly aligned.

\medskip
\noindent \textbf{Advantages.} Locally approximating the RPC model with a perspective camera yields several benefits besides physical interpretability,  efficiency, and simplicity of minimal solvers. First, it enables us to easily convert satellite images for use in existing SfM and MVS pipelines. For instance, Section~\ref{sec:skew_correct} describes how to correct for the skew effect in satellite images. Second, this approximation shows that linear epipolar geometry approximately holds locally in satellite imagery, which avoids the need to handle the complex epipolar lines described by the RPC model as in~\cite{facciolo2017automatic}. This is important for standard vision-based reconstruction pipelines. Third, bundle adjustment is simplified, as demonstrated in Section~\ref{sec:bundle_adjust}, whereas bundle adjusting RPC models is unintuitive and requires expert or proprietary knowledge.

\subsection{Bundle adjustment} \label{sec:bundle_adjust}

Bundle adjustment is a method for refining the parameters of a set of cameras in order to improve their global consistency and accuracy, via minimization of reprojection error of a set of sparse 3D points given their measured projections in the image set~\cite{triggs2000bundle}. %
As shown in prior work~\cite{fraser2005bias,grodecki2003block}, the RPC model is subject to a bias drift in the image domain. We observe that the principal point $(c_x, c_y)$ in the perspective camera model naturally captures any image space translation, and therefore we can bundle adjust just the principal points of all cameras, while keeping the other intrinsic and extrinsic parameters fixed. 
To minimize the distortion caused by projective ambiguity, and to also fix the gauge ambiguity and constrain the reconstruction to the same geographic area, we propose to add a regularization term to the sparse 3D points' coordinates such that they do not drift too far from their original coordinates during bundle adjustment. 
Let $(\mathbf{x}_i, u_i, v_i, \mathbf{p}_{i1}, \mathbf{p}_{i2}, \mathbf{p}_{i3}), i=1,2,\dots, N$ be the 3D-2D correspondences and first three rows of their projection matrices. 
Our proposed regularized bundle adjustment objective is,
\begin{equation}
\sum_{i=1}^N \left(u_i - \frac{\mathbf{p}_{i1}^T \mathbf{x}_i}{\mathbf{p}_{i3}^T \mathbf{x}_i}\right)^2 + \left(v_i - \frac{\mathbf{p}_{i2}^T \mathbf{x}_i}{\mathbf{p}_{i3}^T \mathbf{x}_i}\right)^2 + \lambda \left\Vert \mathbf{x}_i - \hat{\mathbf{x}}_i \right\Vert_2^2,
\end{equation}
where $\hat{\mathbf{x}}_i$ are the point coordinates triangulated with camera parameters before bundle adjustment, $\lambda$ is the tuned regularization weight, $\Vert \cdot \Vert_2$ is the Euclidean norm, and $(\mathbf{x}_i, \mathbf{p}_{i1}, \mathbf{p}_{i2}, \mathbf{p}_{i3}),i=1,2,\dots,N$ are our optimization variables.
We set $\lambda=0.01$ in our experiments.

\subsection{Skew correction}\label{sec:skew_correct}
The skew parameter in an intrinsic matrix is typically assumed to be zero in computer vision approaches, and hence existing MVS pipelines often do not model it. We design a simple skew-correction step to remove skew so that we can use standard pipelines. Our approach decomposes an intrinsic matrix as follows:
\begin{equation}\label{eq:decompose}
\resizebox{0.9\columnwidth}{!} { $
\begin{bmatrix}
    f_x & s & c_x \\
    0 & f_y & c_y \\
    0 & 0 & 1\\
    \end{bmatrix}{=}
        \begin{bmatrix}
    1 & s/f_y & 0 \\
    0 & 1 & 0 \\
    0 & 0 & 1\\
    \end{bmatrix} 
    \begin{bmatrix}
    f_x & 0 & c_x{-}sc_y/f_y\\
    0 & f_y & c_y \\
    0 & 0 & 1\\
    \end{bmatrix}.
$}
\end{equation}
With this decomposition, we then apply the inverse of the first matrix on the right-hand side to transform the images themselves, and use the second matrix on the right-hand side as the new, skew-free intrinsics. 
The resulting image warp resamples the image at the original resolution, and so resolution is preserved by this step. 
Other modifications of intrinsics, such as enforcing non-unit aspect ratio can be performed in a similar way.

%% file: 04-method_mvs.tex
\section{Adapting MVS to satellite imagery}
In this section, we identify and resolve key issues that prevent direct application of standard MVS pipelines tailored for ground-level images to the satellite domain.

\subsection{Depth reparametrization}
MVS pipelines in computer vision often begin by estimating per-view depth maps. Depth is usually defined as the $Z$ component of a 3D point in the reference camera's coordinate frame. Satellite cameras are distant (hundreds or thousands of km) from Earth, leading to a depth distribution with a large mean and comparatively small variance.
This introduces numerical precision issues when  computationally-intensive MVS algorithms are run on consumer GPUs with only single-precision floating point, which supports only 7 effective decimal digits.

We tackle this challenge by reparametrizing depth using plane-plus-parallax~\cite{irani1998from}: we first choose a plane close to the scene and parallel to the ground plane as our reference plane. Then the distance between a 3D point and this reference plane is defined as the reparametrized depth, which is usually bounded by hundreds of meters. Concretely, this amounts to adding a fourth row to the 3 by 4 projection matrix:
\begin{equation}\label{eq:depth_reparam}
\resizebox{0.8\columnwidth}{!}{$
    \begin{bmatrix}
    u \\
    v\\
    1\\
    m
    \end{bmatrix}=
    \begin{bmatrix}
    uZ \\
    vZ \\
    Z \\
    mZ
    \end{bmatrix}=
    \begin{bmatrix}
    \mathbf{P}_{11} & \mathbf{P}_{12} & \mathbf{P}_{13} & \mathbf{P}_{14} \\
    \mathbf{P}_{21} & \mathbf{P}_{22} & \mathbf{P}_{23} & \mathbf{P}_{24} \\
    \mathbf{P}_{31} & \mathbf{P}_{32} & \mathbf{P}_{33} & \mathbf{P}_{34} \\
    0   & 0 & \bar{Z} & -\bar{Z}d 
    \end{bmatrix}
    \begin{bmatrix}
    x \\
    y \\
    z \\
    1
    \end{bmatrix},
$}
\end{equation}
where $Z=\mathbf{P}_{31}x+\mathbf{P}_{32}y+\mathbf{P}_{33}z+\mathbf{P}_{34}$ is the conventional depth; the plane $(0, 0, 1, d)$\footnote{We denote the plane equation $\mathbf{n}^Tx-d=0$ as $(\mathbf{n}, d)$ for convenience.} is chosen to lie below the scene, i.e., $z>d$ holds for all scene points; $\bar{Z}$ is the average conventional depth of all the sparse scene points in the SfM step. With these careful choices, the reparametrized depth $m=(0\cdot x+0\cdot y + \bar{Z}\cdot z - \bar{Z}\cdot d)/Z\approx z-d$ is a positive variable generally bounded by hundreds of meters.\footnote{The second equality is due to the fact that $Z$ has a large mean but small variance so that $\bar{Z}/Z\approx 1$; $z-d$ is the distance between the the point $(x,y,z)$ and the plane $(0,0,1,d)$.} The augmented $4 \times 4$ projection matrix is finally scaled such that the largest entry is 1. 

\subsection{Stable homography computation}
To measure photo-consistency between two views, many modern MVS systems use homographies to relate an image patch in a reference view to one in a source view. The homography matrix is computed as $\mathbf{H}=\mathbf{K}_2\left(\mathbf{R}_{12}-\frac{\mathbf{n}^T \mathbf{t}_{12}}{d}\right)\mathbf{K}_1^{-1}$,
where $(\mathbf{n}, d)$ is a plane in camera one's coordinate frame, and $\mathbf{R}_{12}, \mathbf{t}_{12}$ is camera two's relative pose with respect to camera one. We find that this expression for computing a homography can be numerically unstable under single-precision, due to large intermediate values. To address this issue, we derive an alternative way to compute the homography directly from the $4 \times 4$ projection matrix in Eq.~\ref{eq:depth_reparam} without involving large numbers.
 For convenience, denote pixel coordinates as $\mathbf{u}=[u;v;1]$, scene coordinates as $\mathbf{x}=[x;y;z;1]$, and projection matrices (augmented with a fourth row) as $\mathbf{P}\in \mathcal{R}^{4\times 4}$. Given a plane $[\mathbf{n};-d]^T\mathbf{x}=0$, and the projection equation of two cameras $[\mathbf{u}_1;m_1]=\mathbf{P}_1\mathbf{x}, [\mathbf{u}_2;m_2]=\mathbf{P}_2\mathbf{x}$,
our goal is to derive a homograhy matrix $\mathbf{H}\in \mathcal{R}^{3\times 3}$ such that $\mathbf{u}_2=\mathbf{H}\mathbf{u}_1$.

\medskip
\noindent \textbf{Derivation.} 
The plane equation implies that
\begin{equation}
    [\mathbf{n};-d]^T\mathbf{x}=[\mathbf{n};-d]^T\mathbf{P}_1^{-1}[\mathbf{u}_1;m_1]=0.
\end{equation}
Let us write the $1 \times 4$ vector $[\mathbf{n};-d]^T\mathbf{P}_1^{-1}$ as $[\mathbf{q}^T,r]$, with $\mathbf{q}\in\mathcal{R}^3, r\in \mathcal{R}$. Then $\mathbf{q}^T\mathbf{u}_1+rm_1=0$, i.e., $m_1=-\frac{\mathbf{q}^T\mathbf{u}_1}{r}$. Hence we have
\begin{align}
    [\mathbf{u}_2;m_2]&=\mathbf{P}_2\mathbf{P}_1^{-1}[\mathbf{u}_1;m_1]
    =\mathbf{P}_2\mathbf{P}_1^{-1}[\mathbf{u}_1;-\frac{\mathbf{q}^T\mathbf{u}_1}{r}] \nonumber \\
    &=\mathbf{P}_2\mathbf{P}_1^{-1}[\mathbf{I};-\frac{\mathbf{q}^T}{r}]\mathbf{u}_1, \label{eq:homo}
\end{align}
where $\mathbf{I}\in \mathcal{R}^{3\times 3}$ is an identity matrix. Let us define $\mathbf{A}=\mathbf{P}_2\mathbf{P}_1^{-1}[\mathbf{I};-\frac{\mathbf{q}^T}{r}]$. Then the homography matrix is $\mathbf{H}=\mathbf{A}_{1:3, 1:3}$. One can easily check that Eq.~\ref{eq:homo} involves numbers with limited magnitude, given that $\mathbf{P}_{1,2}$ are pre-scaled. Analysis of its numerical stability can be found in the supplemental document.

\subsection{Plane-sweep stereo}
In addition to integrating the depth re-parametrization and stable homography computation methods into MVS in COLMAP (which is based on PatchMatch stereo~\cite{bleyer2011patchmatch}), we also evaluate the use of standard plane-sweep stereo, with ground-parallel planes, as an alternate MVS module. This usage is motivated by the observation that the fronto-parallel assumption often holds very strongly for urban scenes if we sweep planes parallel to the ground.
We adopt the census cost as a photo-consistency measure, and compute a cost volume. We further smooth each cost slice using a guided filter as in \cite{hosni2012fast}. Plane-sweep and cost volume filtering are very efficient; if accuracy is not a top concern, then the arg minimum at each pixel of the filtered cost volume can serve as the height map. However, we can also compute a refined height map using Markov Random Field (MRF) optimization techniques, which are especially helpful in areas of weak texture, as we demonstrate in our experiments.

%% file: 05-experiments.tex
\section{Experiments}
In this section, we provide empirical evidence to demonstrate the validity of our proposed adapted pipeline on a benchmark dataset we created from MVS3DM dataset~\cite{bosch2016multiple}. We first evaluate the SfM module, and then the MVS module.

\subsection{Dataset and metrics}

We tested our methods on three sites from the IARPA MVS3DM dataset~\cite{bosch2016multiple}. Tab.~\ref{tab:dataset} summarizes the data. The images were captured over a span of two years by the WorldView-3 satellite with a resolution of 30cm per pixel in nadir views. 3D ground-truth is captured by airborne lidar. 

\begin{table}[t]
    \caption{Overview of the data. Avg.\ \# Pix stands for average number of pixels over all views.}
    \centering
    \begin{tabular}{lccc}
          \toprule[1pt]
            & \# Views & Avg. \# Pix & Area ($\text{km}^2$) \\
           \midrule
        site 1 & 47 & 4.22M & 0.464\\
        site 2  &47 & 1.27M & 0.138\\
        site 3 & 40 & 1.44M & 0.150\\
        \bottomrule[1pt]
    \end{tabular}
\label{tab:dataset}
\end{table}
During evaluation, we follow previous practice~\cite{bosch2016multiple} and flatten the ground-truth point cloud and the reconstructed one onto the same pre-specified geographic grid to obtain two height maps. Each geographic grid cell is $0.5\times 0.5\text{m}^2$. The maps are then aligned and compared pixel-wise to compute metric scores, namely median height error and completeness (percentage of points with error less than a threshold of 1m).

\subsection{Structure from motion}
We evaluate our perspective camera approximation using three metrics: accuracy of forward projection, accuracy of triangulation, and consistency between our bundle-adjusted perspective cameras and S2P's pointing-error-corrected RPC models. For concision, the following experiments are done on site 1, but the conclusions generalize to other sites.

As mentioned above, our approximated perspective cameras are derived using the $M^3$ ($M=100$) grid points sampled in the bounding volume. 
The average of per-view maximum forward projection error compared with the RPC model on these grid points is 0.194px. Further, we apply both the RPC model and the approximate perspective camera to points in the ground-truth point cloud. The average of per-view maximum forward projection error is 0.126px, which leads to as small as a maximum 3.8cm loss of accuracy given an image resolution of 30cm per pixel.

\begin{figure}[t]
    \centering
\begin{minipage}{\columnwidth}
\centering
    \includegraphics[width=0.85\columnwidth]{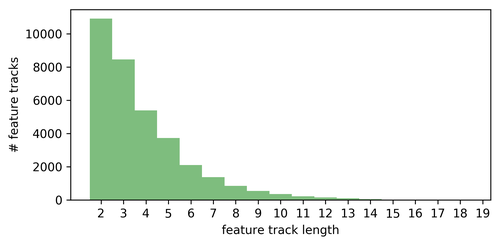}
    \caption{Feature track length, site 1 (average length is 3.85).}
    \label{fig:track_length}
\end{minipage}
\begin{minipage}{\columnwidth}
\centering
    \includegraphics[width=0.85\columnwidth]{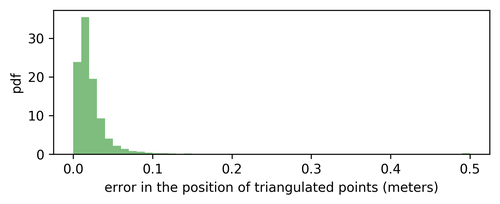}
    \caption{Triangulation error
    of approximate perspective camera with respect to the RPC model for site 1.}
    \label{fig:inv_triangulate_err}
\end{minipage}
\end{figure}

Second, we evaluate the triangulation accuracy of our perspective cameras. We first run standard feature detection and matching across all views, then triangulate the resulting feature tracks using both the approximate perspective and RPC models. The distribution of the number of views across $\sim$34K
feature tracks is shown in Fig.~\ref{fig:track_length}. Perspective camera triangulation is performed with COLMAP, and for RPC triangulation we implemented a solver based on~\cite{rpc_3d}. We then compute the difference in point locations between the perspective- and RPC-triangulated 3D points. 
The distribution of differences is shown in Fig.~\ref{fig:inv_triangulate_err}; the majority of the differences are below 5cm, with a few large differences caused by incorrect correspondences. 

\begin{figure}[t]
\centering
  \includegraphics[width=0.85\columnwidth]{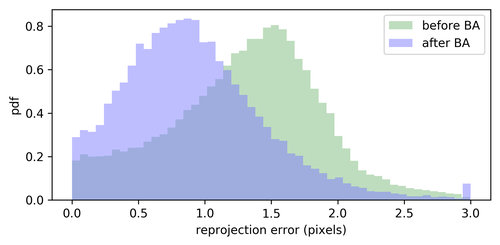}
  \caption{Distribution of reprojection errors before and after bundle adjustment, site 1.}
  \label{fig:reproj_err}
\end{figure}

\begin{figure}[t]
\centering
\begin{minipage}{\columnwidth}
    \begin{tabular}{c@{\hspace{0.1em}}c@{\hspace{0.1em}}c@{\hspace{0.1em}}}
            {\small RPC } & {\small Perspective} & {\small Error} \\
         \includegraphics[valign=c,width=0.33\columnwidth]{{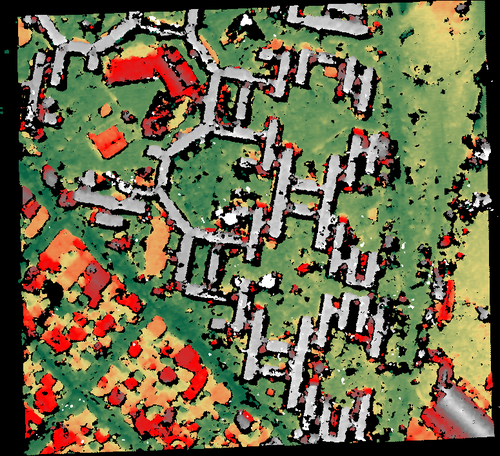}} & 
         \includegraphics[valign=c,width=0.33\columnwidth]{{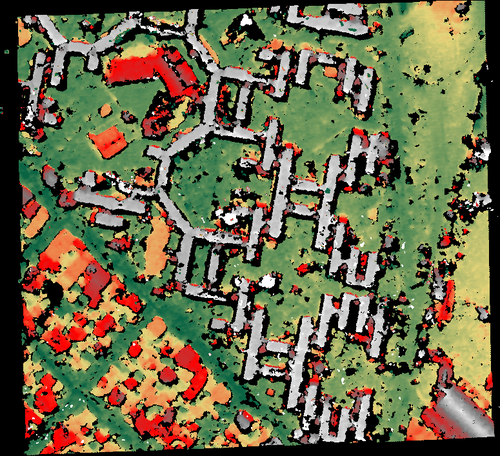}} & 
         \includegraphics[valign=c,width=0.33\columnwidth]{{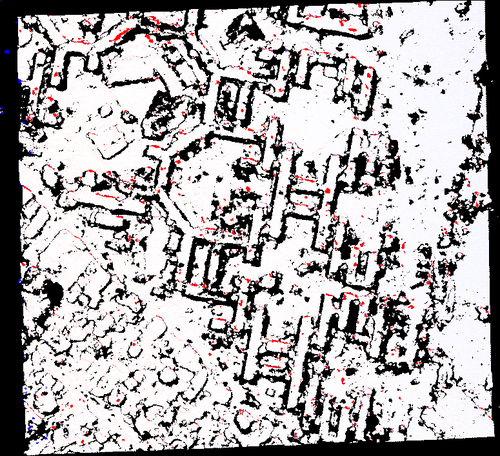}} 
     \end{tabular} 
  \caption{Consistency between S2P's point-error-corrected RPC and our bundle-adjusted perspective camera.
  }\label{fig:consistency}
\end{minipage}

\begin{minipage}{\columnwidth}
\centering
\begin{tabular}{cc}
     \includegraphics[width=0.49\columnwidth]{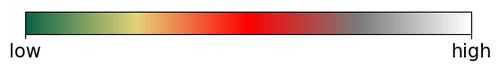} &
      \includegraphics[width=0.49\columnwidth]{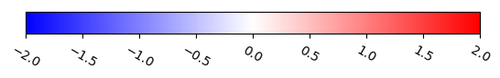}
\end{tabular}\\
    \caption{Color bars used for displaying height maps and error maps in this paper; units are meters.}
    \label{fig:color_bar}
\end{minipage}
\end{figure}

Third,
we investigate the effect of bundle adjustment. 
The median reprojection error over $\sim$34K
sparse points reduces from 1.36px to 0.864px after bundle adjustment; the distribution of reprojection errors is shown in Fig.~\ref{fig:reproj_err}. This reduction in reprojection error indicates better consistency among cameras and final reconstruction accuracy because our bundle adjustment scheme is free from projective ambiguity. We also compare the bundle-adjusted perspective cameras to the RPC model whose pointing error is corrected by S2P. 
We take the correspondence map between two views computed by S2P, and re-triangulate with our bundle-adjusted perspective cameras. We then compare our triangulated point cloud to S2P's point cloud triangulated with RPC model using the evaluation code. The median height difference is just 2.72cm. The error map is shown in Fig.~\ref{fig:consistency}.

The above three aspects demonstrate the high accuracy of the approximated perspective camera for the RPC model in a local area. The entire SfM procedure takes about 1.68 minutes\footnote{All runtimes are reported on a machine with 16 CPUs, 60GB memory, and 3 GPUs.} for this site with 46 images, which is very efficient thanks to the fact that the SfM pipeline was initially designed to handle large-scale Internet photo collections.

\begin{figure}[t]
\centering
    \begin{tabular}{c@{\hspace{0.1em}}c@{\hspace{0.1em}}}
         {\small S2P (CP=64.2\%)} & {\small COLMAP (CP= 61.2\%)}\\
         \includegraphics[valign=c,width=0.42\columnwidth]{{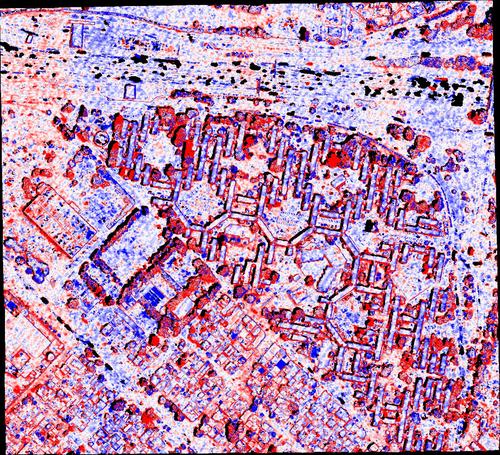}} &
         \includegraphics[valign=c,width=0.42\columnwidth]{{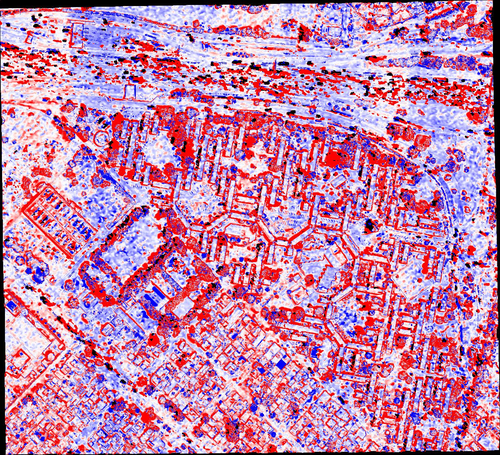}}\\
         {\small PSS+CVF (CP= 67.5\%)} & {\small PSS+CVF+MRF (CP=68.3\%)}\\
         \includegraphics[valign=c,width=0.42\columnwidth]{{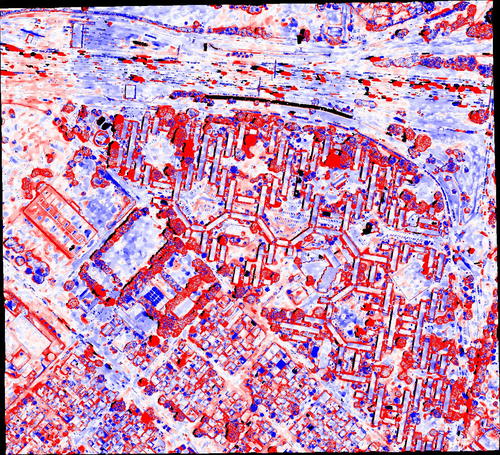}} &
         \includegraphics[valign=c,width=0.42\columnwidth]{{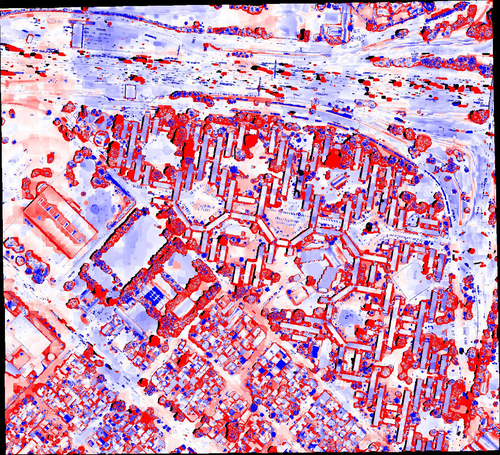}}
     \end{tabular} 
  \caption{Error maps of different algorithms on the same reference and source views (captured seconds apart).
  }\label{fig:compare_two_view}
\end{figure}

\begin{figure*}[!htb]
\centering
    \begin{tabular}{l@{\hspace{0.1em}}c@{\hspace{0.1em}}c@{\hspace{0.1em}}c@{\hspace{0.1em}}c@{\hspace{0.1em}}}
         &{\small Example View} & {\small Lidar GT } & {\small COLMAP} & {\small Error} \\
        {\small Site 1  } & \includegraphics[valign=c,width=0.205\textwidth]{{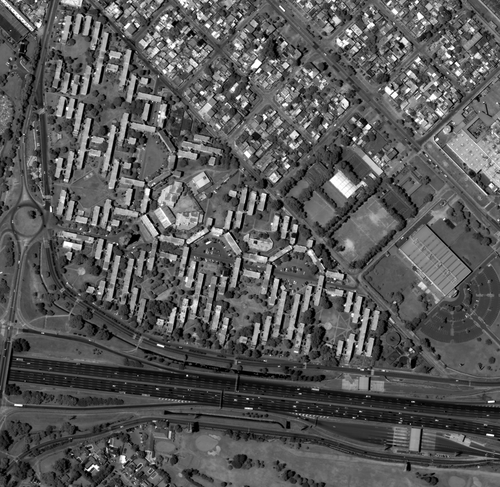}} & 
         \includegraphics[valign=c,width=0.22\textwidth]{{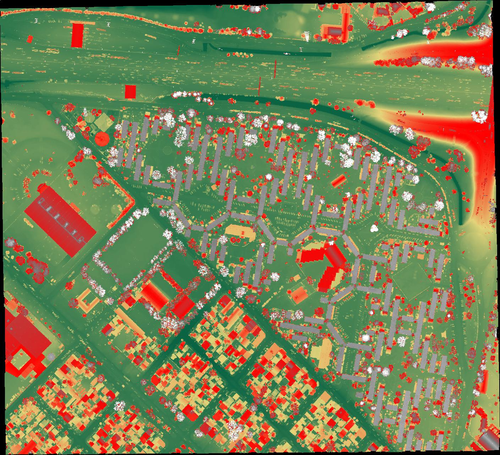}} & 
         \includegraphics[valign=c,width=0.22\textwidth]{{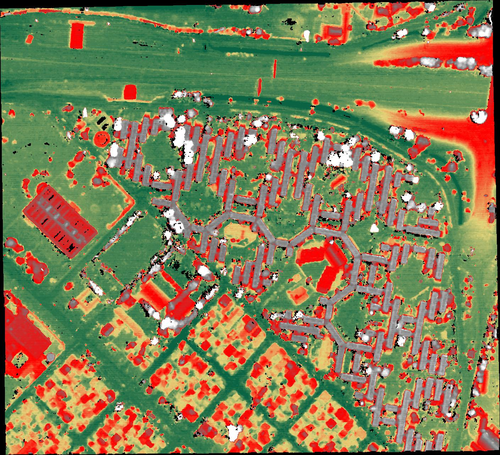}} & 
         \includegraphics[valign=c,width=0.22\textwidth]{{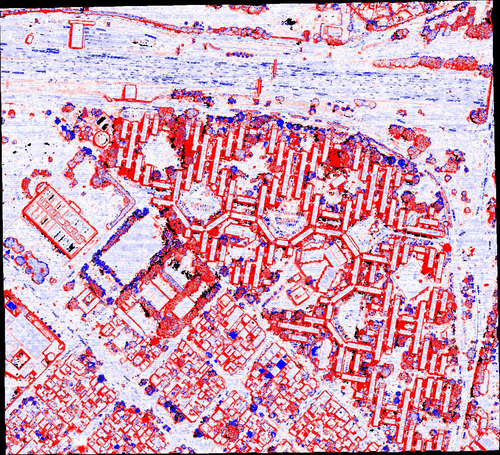}} \\[1.4cm]
         {\small Site 2~~~} & \includegraphics[valign=c,width=0.205\textwidth]{{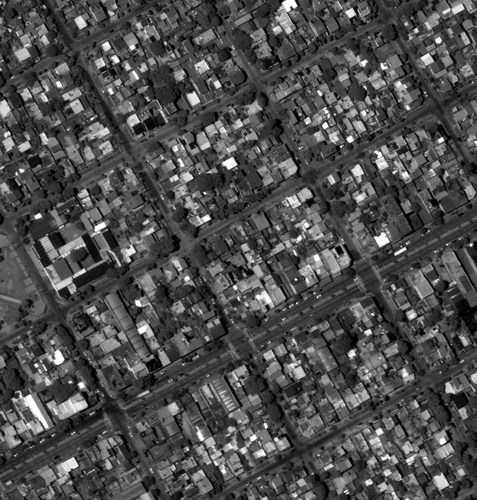}} & 
         \includegraphics[valign=c,width=0.22\textwidth]{{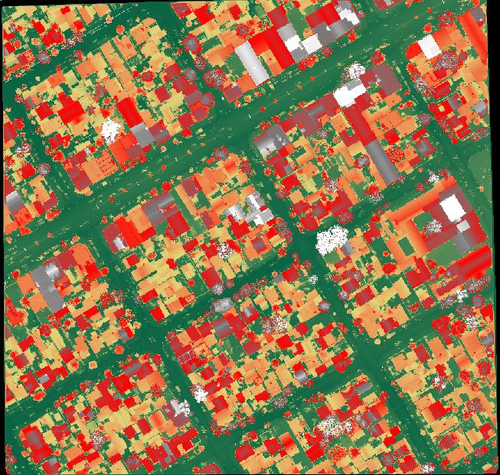}} & 
         \includegraphics[valign=c,width=0.22\textwidth]{{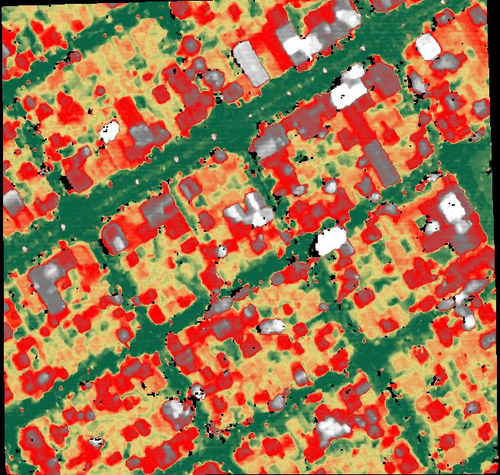}} & 
         \includegraphics[valign=c,width=0.22\textwidth]{{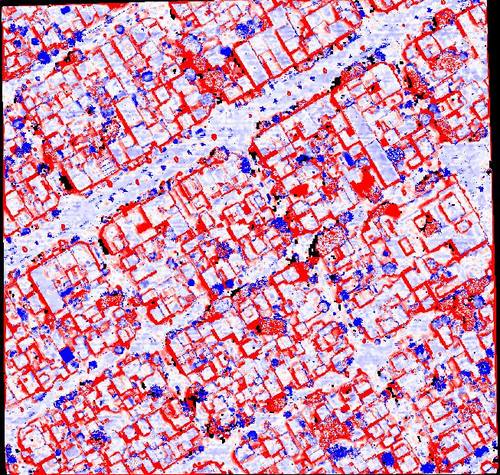}} \\[1.4cm]
         {\small Site 3} & \includegraphics[valign=c,width=0.205\textwidth]{{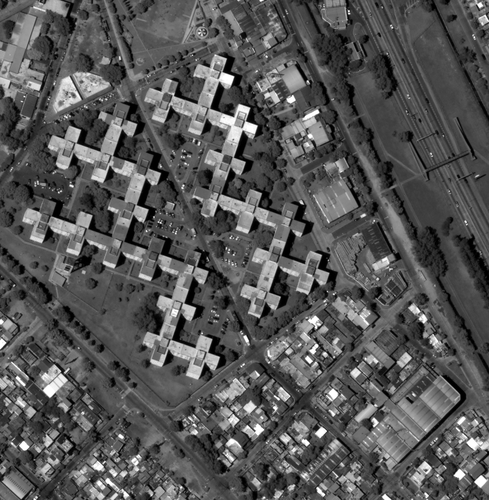}} & 
         \includegraphics[valign=c,width=0.22\textwidth]{{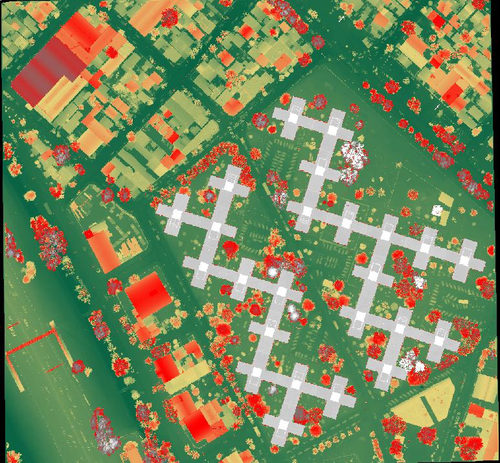}} & 
         \includegraphics[valign=c,width=0.22\textwidth]{{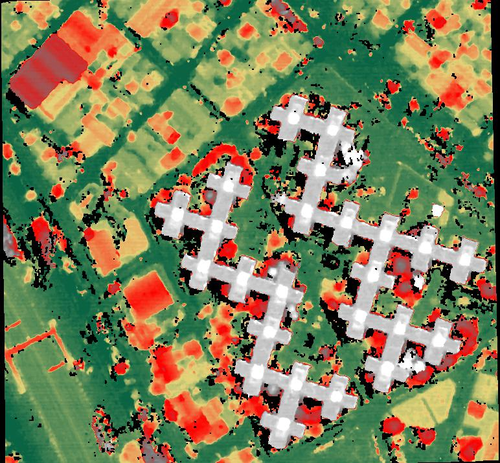}} & 
         \includegraphics[valign=c,width=0.22\textwidth]{{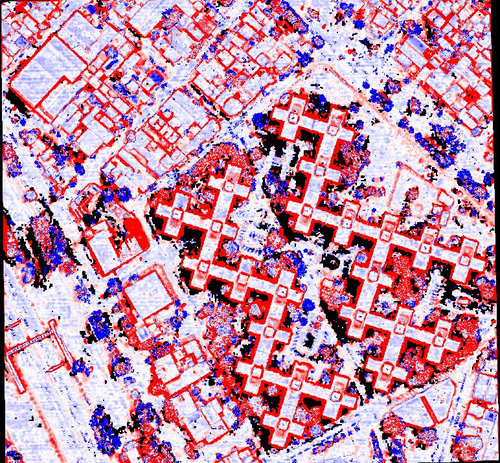}} \\
     \end{tabular} 
  \caption{Qualitative results of COLMAP MVS. Overall, the reconstruction is very good, despite some room for improvement at building boundaries and weakly-textured regions (e.g., ground and building tops). 
  }\label{fig:main_qualitative}
\end{figure*}

\subsection{Multi-view stereo}

We now evaluate our adapted MVS module by running: 1) COLMAP's MVS enhanced with our proposed depth re-parametrization and stable homography computation, 2) different methods including plane-sweep stereo compared on stereo pairs.

We report the accuracy of our enhanced COLMAP MVS run on \emph{all} images in each site on sites 1-3 in Tab.~\ref{tab:main_result}. Qualitative results are shown in Fig.~\ref{fig:main_qualitative}. Considering COLMAP MVS's greater efficiency and scalability, these results are quite good, despite the fact that the accuracy is lower than the S2P-based method~\cite{de2014automatic2}. 
We believe this gap in accuracy is mainly due to the fact that COLMAP MVS's core stereo component is efficiency-driven.
In particular, it sometimes struggles at object boundaries and in weakly-textured regions, as can be seen from the errors surrounding buildings, on the ground, and on building tops. Similar observations are also reported recently on ground-level images~\cite{schops2017multi}. However, we also note that an interesting aspect of this experiment is that COLMAP is leveraging all images at once, which is quite distinct from the practice of S2P, which only considers pairs of images at a time. We believe that these results are promising and suggest that further explorations of diverse handling of images in the satellite domain will be fruitful.

Next, we 
compare different algorithms on the minimal setting of stereo pairs.
In particular, we compare the standard S2P to COLMAP configured to run on just two views, as well as various settings of plane-sweep stereo. For each site, we select 10 image pairs on which the baseline S2P performs very well; all algorithms are tested on these same image pairs, so that the comparison focuses on stereo accuracy, isolated from the influence of view selection and aggregation. The average metric performance over the selected pairs is shown in Tab.~\ref{tab:compare_two_view}. The experiments show that COLMAP MVS is the least accurate while being the fastest, while our PSS+CVF+MRF\footnote{We use graph cuts in these experiments~\cite{boykov1999fast, boykov2004experimental}; the system can be further accelerated with faster MRF solvers, e.g., SGM-like solvers~\cite{hirschmuller2007stereo, facciolo2015mgm}.} outputs comparable results to S2P. Furthermore, we found that both COLMAP and PSS+CVF, which lack global optimization, suffer from the errors on weakly-textured areas, e.g., ground and large building tops, while MRF helps reduce the errors (see Fig.~\ref{fig:compare_two_view}).

\begin{table}[t]
    \caption{Metric result of COLMAP MVS;  CP stands for completeness, ME for median height error.}
    \centering
    \begin{tabular}{lccc}
          \toprule[1pt]
            &   CP (\%) & ME (m) & Time (mins)\\
            \midrule
        site 1 & 72.5 & 0.315 & 18.7 \\
        site 2 & {66.8} & {0.450} & 7.37 \\
        site 3 & {63.4} & {0.393} & 6.94\\
        \bottomrule[1pt]
    \end{tabular}
\label{tab:main_result}
\end{table}

\begin{table}[ht]
    \caption{Comparison between different algorithms on stereo pairs; for each site, both metrics and time are averaged over 10 selected pairs on which S2P performs very well. PSS: plane sweep stereo. CVF: cost volume filtering. MRF: Markov random field.  }
    \centering
    \resizebox{\columnwidth}{!}{
    \begin{tabular}{llccc}
          \toprule[1pt]
        &         &  CP (\%) & ME (m) & Time (mins)\\
        \midrule
        Site 1 & S2P & 65.6 & 0.432 & 16.8\\
         & COLMAP& 58.0  &  0.648 & 0.374 \\
        &PSS+CVF&  68.1 & 0.442 & 4.40 \\
        &PSS+CVF+MRF&  \textbf{69.1} & \textbf{0.395}  & 29.7\\
        \midrule %
        Site 2 & S2P & 62.8 & \textbf{0.435} & 2.63\\
            & COLMAP & 59.3 & 0.689 & 0.196\\
            & PSS+CVF & 63.8 & 0.502 & 1.19\\
            & PSS+CVF+MRF & \textbf{64.0} & 0.478 & 6.75\\
        \midrule %
        Site 3 & S2P & 53.9 & \textbf{0.421} & 2.78 \\
              &    COLMAP & 42.1 & 0.949 & 0.208\\
             & PSS+CVF & 57.1 & 0.591 & 2.39 \\
             & PSS+CVF+MRF & \textbf{58.8} & 0.521 & 14.8\\
        \bottomrule[1pt]
    \end{tabular}
    }
    \label{tab:compare_two_view}
\end{table}

%% file: 06-conclusion.tex
\section{Conclusion}
In this work, we propose a set of mechanisms to adapt the satellite-image-based 3D reconstruction problem to one that can be solved with computer vision pipelines designed for ground-level images. We show the effectiveness of the proposed methods by incorporating them into COLMAP, a state-of-the-art SfM and MVS system. We evaluate this pipeline, as well as plane-sweep stereo, on a remote sensing dataset. 
We will publish our code to encourage future research in brigding the gap between satellite stereo and vision reconstruction.